\documentclass[10pt,conference,letterpaper]{IEEEtran}
\pdfoutput=1

\ifCLASSINFOpdf
\else
\fi
\usepackage[utf8]{inputenc} 
\usepackage[T1]{fontenc}    
\usepackage{hyperref}       
\usepackage{url}            
\usepackage{booktabs}       
\usepackage{amsfonts}       
\usepackage{nicefrac}       
\usepackage{microtype}      
\usepackage{etex}
\usepackage[square,numbers]{natbib}
\usepackage[cmex10]{amsmath}
\usepackage{tikz}
\usepackage{array}
\usepackage{ragged2e}
\usepackage{tabularx}
\usepackage{subcaption}
\usepackage[font=footnotesize]{caption}
\usepackage{graphicx}
\graphicspath{{charts/}{pics/}{figs/}}
\DeclareGraphicsExtensions{.pdf,.jpeg,.png}
\usepackage{pgfplots}
\usepackage[export]{adjustbox}
\usepackage{calc}
\pgfplotsset{compat=newest} 
\pgfplotsset{plot coordinates/math parser=false} 
\usepackage{multirow}
\usepackage{xcolor}

\usepackage{fancyhdr}
\usepackage{kantlipsum}

\fancypagestyle{plain}{
	\fancyhf{}
	\fancyhead[C]{Conference on \LaTeX}     
	\fancyfoot[L]{This is a notice}

}
\usepackage{eso-pic}

\RequirePackage{natbib}

\begin{document}
%

\title{Recursive Autoconvolution for Unsupervised Learning of Convolutional Neural Networks}

\author{\IEEEauthorblockN{Boris Knyazev$^{*}$\textsuperscript{†}, Erhardt Barth\textsuperscript{†} and Thomas Martinetz\textsuperscript{†}}
\IEEEauthorblockA{$^{*}$Bauman Moscow State Technical University\\
Email: \texttt{bknyazev@bmstu.ru, borknyaz@gmail.com}}
\IEEEauthorblockA{\textsuperscript{†}Institut für Neuro- und Bioinformatik, University of Lübeck\\
Email: \texttt{\{barth, martinetz\}@inb.uni-luebeck.de}
}}

\maketitle
\newcommand{\mSize}{2.0pt}
\newcommand{\lWidth}{0.8pt}
\newcommand{\imSize}{34pt}
\newcommand{\rangeConvOneThree}{1,3}
\newcommand{\rangeConvZeroThree}{0,3}
\newcommand{\rangeConvOneTwo}{1,2}
\newcommand{\rangeConvZeroTwo}{0,2}
\newcommand{\scaleFilters}{0.54}
\definecolor{outdatedResultsColor}{rgb}{0.00000,0.3,0.7}%
\newcommand{\outdatedResults}[1]{\textit{\textcolor{outdatedResultsColor}{#1}}}

\newcommand{\argmin}{\operatornamewithlimits{arg\,min}}
\newcommand{\comment}[1]{}

\begin{abstract}
In visual recognition tasks, such as image classification, unsupervised learning exploits cheap unlabeled data and can help to solve these tasks more efficiently.
We show that the recursive autoconvolution operator, adopted from physics, boosts existing unsupervised methods by learning more discriminative filters.
We take well established convolutional neural networks and train their filters layer-wise. In addition, based on previous works we design a network which extracts more than 600k features per sample, but with the total number of trainable parameters greatly reduced by introducing shared filters in higher layers. We evaluate our networks on the MNIST, CIFAR-10, CIFAR-100 and STL-10 image classification benchmarks and report several state of the art results among other unsupervised methods.
\end{abstract}


%
\IEEEpeerreviewmaketitle

\section{Introduction}
Large-scale visual tasks can now be solved with big deep neural networks, if thousands of labeled samples are available and if training time is not an issue. Efficient GPU implementations of standard computational blocks make training and testing feasible. 

A major drawback of supervised neural networks is that they heavily rely on labeled data. It is true that in real applications it does not really matter which methods are used to achieve the desired outcome. But in some cases,
labeling can be an expensive process. 
However, visual data are full of abstract features unrelated to object classes. Unsupervised learning exploits abundant amounts of these cheap unlabeled data and can help to solve the same tasks more efficiently. In general, unsupervised learning is important for moving towards artificial intelligence \citep{lecun2015deep}.

In this work, we learn a visual representation model for image classification, which is particularly effective (in terms of accuracy) when the number of labels is relatively small. As a result, our model can potentially be successfully applied to other tasks (e.g., biomedical data analysis or anomaly detection), in which label information is often scarce or expensive.

We are inspired by the previous works in which network filters (or weights) are learned layer-wise without label information
\citep{labusch2008simple,coates2011analysis,coates2011importance,le2011ica,dundar2015convolutional,miclut2014committees,lin2014stable}.
In accordance with these works, we thoroughly validate our models and confirm its advantage in the tasks with only few labeled training samples, such as STL-10 \citep{coates2011analysis} and reduced variants of MNIST \citep{lecun1998gradient} and CIFAR-10 \citep{krizhevsky2009learning}.
In addition, on full variants of these datasets and on CIFAR-100 we demonstrate that unsupervised learning is steadily approaching performance levels of supervised models, including convolutional neural networks (CNNs) trained by backpropagation on thousands of labeled samples \citep{zeiler2013stochastic, mairal2014convolutional, zhao2015stacked}.
This way, we provide further evidence that unsupervised learning is promising for building efficient visual representations.

\begin{figure}[t]
	\hspace{-0.7em}
	\setlength\fboxsep{0pt}
	\setlength{\fboxrule}{1pt}
	\begin{minipage}[c][3.8cm][t]{0.5\textwidth}
		\centering
		\captionsetup[subfigure]{labelformat=empty, font=small,skip=2pt}
		\subcaptionbox{(a)\label{sub-mnist}}{
			\captionsetup[subfigure]{labelformat=empty, font=footnotesize, position=top,skip=-1pt}
			\subcaptionbox{$n=0$}{%
				\includegraphics[height=\imSize]{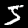}%
			}
			\subcaptionbox{$n=1$}{%
				\fcolorbox{green}{green}{\includegraphics[height=\imSize]{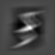}}%
			}
			\subcaptionbox{$n=2$}{%
				\fcolorbox{green}{green}{\includegraphics[height=\imSize]{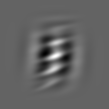}}%
			}
		}
		\subcaptionbox{(b)\label{sub-cifar10}}{
			\captionsetup[subfigure]{labelformat=empty, font=footnotesize, position=top,skip=-1pt}
			\subcaptionbox{$n=0$}{%
				\includegraphics[height=\imSize]{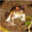}%
			}
			\subcaptionbox{$n=1$}{%
				\fcolorbox{green}{green}{\includegraphics[height=\imSize]{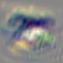}}%
			}
			\subcaptionbox{$n=2$}{%
				\fcolorbox{green}{green}{\includegraphics[height=\imSize]{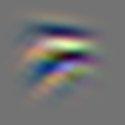}}%
			}
		}
		\par\medskip
		\vspace{-0.1cm}
		\subcaptionbox{(c)\label{sub-stl10}}{
			\captionsetup[subfigure]{labelformat=empty, font=footnotesize, position=top,skip=-1pt}
			\subcaptionbox{$n=0$}{%
				\includegraphics[height=\imSize]{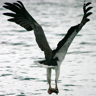}%
			}
			\subcaptionbox{$n=1$}{%
				\fcolorbox{green}{green}{\includegraphics[height=\imSize]{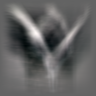}}%
			}
			\subcaptionbox{$n=2$}{%
				\fcolorbox{green}{green}{\includegraphics[height=\imSize]{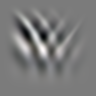}}%
			}
		}
		\subcaptionbox{(d)\label{sub-stl10-2}}{
			\captionsetup[subfigure]{labelformat=empty, font=footnotesize, position=top,skip=-1pt}
			\subcaptionbox{$n=0$}{%
				\includegraphics[height=\imSize]{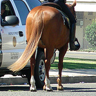}%
			}
			\subcaptionbox{$n=1$}{%
				\fcolorbox{green}{green}{\includegraphics[height=\imSize]{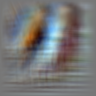}}%
			}
			\subcaptionbox{$n=2$}{%
				\fcolorbox{green}{green}{\includegraphics[height=\imSize]{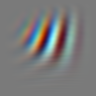}}%
			}
		}
	\end{minipage}%
	\vspace{0.5em}
	\caption{
		{Recursive autoconvolution (Eq. \ref{eq_rec_autoconv_2}) of orders $n=0,1,2$ applied to samples: MNIST (a); CIFAR-10 (b); STL-10 (c,d). Note how the patterns obtained for $n=1,2$ (highlighted in green frames) are similar to the low and mid-level features of CNNs (e.g., visualized in Fig. 2 in \citep{zeiler2014visualizing}). The novelty of this work is that we learn filters from these patterns.}  
	}
	\vspace{-1em}
	\label{fig_main}
\end{figure}

The main contribution of this work is adaptation of the recursive autoconvolution operator \citep{knyazev2014convolutional} for convolutional architectures. 
Concretely, we demonstrate that this operator can be used together with existing clustering methods (e.g., $k$-means) or other learning methods (e.g., independent component analysis (ICA)) to train filters that resemble the ones learned by
CNNs and, consequently, are more discriminative (Fig. \ref{fig_main}, \ref{fig_main2} and Sections \ref{section_autoconv}, \ref{section_learning}).
Secondly, we substantially reduce the total number of learned filters in higher layers of some networks without loss of classification accuracy (Section \ref{section:shared_filters}), which allows us to train larger models (such as our AutoCNN-L32 with over 600k features in the output).
Finally, we report several state of the art results among unsupervised methods while keeping computational cost relatively low (Section \ref{section:large_exper}).

\section{Related work}
Unsupervised learning is used quite often 
as an additional regularizer in the form of weights initialization \citep{paine2014analysis}
or reconstruction cost \citep{zhao2015stacked,rasmus2015semi}, or as an independent visual model trained on still images \citep{dosovitskiy2014discriminativeLatest, krizhevsky2009learning} or image sequences \citep{wang2015unsupervised}.

However, to a larger extent, our method is related to another series of works
\citep{labusch2008simple,coates2011analysis,coates2011importance,le2011ica,hui2013direct,lin2014stable}
and, in particular, \citep{dundar2015convolutional,miclut2014committees}, in which filters (or some basis) are learned layer-wise and, contrary to the methods above, neither backpropagation nor fine tuning is used.

In these works, learning filters with clustering methods, such as $k$-means, is a standard approach \citep{coates2011analysis, hui2013direct, lin2014stable, miclut2014committees, dundar2015convolutional}. For this reason, and to make comparison of our results easier, $k$-means is also adopted in this work as a default method. Moreover, clustering methods can learn overcomplete dictionaries without additional modifications, such as done for ICA in \citep{le2011ica}. Nevertheless, since ICA \citep{hyvarinen1999fast} is also a common practice to learn filters
, we conduct a couple of simple experiments with this method, as well as with principal component analysis (PCA), to probe our novel idea more thoroughly.
In contrast to various popular coding schemes \citep{labusch2008simple, coates2011analysis, coates2011importance, lin2014stable},
our forward pass is built upon a well established supervised method - a convolutional neural network \citep{lecun1998gradient}. 

Recently, convolutional networks were successfully trained layer-wise in an unsupervised way \citep{dundar2015convolutional, miclut2014committees}.
In their works, as well as in our work, the forward pass is mostly kept standard, while methods to learn stronger (in terms of classification) filters are developed. For instance, in \citep{dundar2015convolutional}, $k$-means is enhanced by introducing convolutional clustering. Convolutional extension of clustering and coding methods is one of the ways to reduce redundancy in filters and improve classification, e.g., convolutional sparse coding \citep{bristow2013fast}. In this work, we suggest another concept of making filters more powerful, namely, by recursive autoconvolution applied to image patches before unsupervised learning.

Autoconvolution (or self-convolution) and its properties seem to have been first analyzed in physics (in spectroscopy \citep{dose1981inversion}) and later in function optimization \citep{gorenflo1994autoconvolution} as the problem of deautoconvolution arose.
This operator also appeared in visual tasks to extract invariant patterns \citep{heikkila2002multi}. But, to the best of our knowledge, its recursive version, used as a pillar in this work, was first suggested in \citep{knyazev2014convolutional} for parametric description of images and temporal sequences. By contrast, we use this operator to learn convolution kernels in a multilayer CNN, which we refer to as an AutoCNN.

\section{Autoconvolution}
\label{section_autoconv}

Our filter learning procedure, discussed further in Section \ref{section_learning}, is largely based on autoconvolution and its recursive extension. In this and the next sections, we formulate the basics of these operators in general terms.

We first describe the routine for processing arbitrary discrete data based on autoconvolution. 
It is convenient to consider autoconvolution in the frequency domain. According to the convolution theorem, for $N$-dimensional discrete signals $ \mathbf{X} $ and $ \mathbf{Y} $, such as images ($N=2$): 
$ \mathcal{F}(\mathbf{X} \ast \mathbf{Y}) = k\mathcal{F}(\mathbf{X}) \circ \mathcal{F}(\mathbf{Y}) $, where $\mathcal{F}$ - the $N$-dimensional forward discrete Fourier transform (DFT), $\circ$ - point-wise matrix product, $k$ - a normalizing coefficient (which will be ignored further, since we apply normalization afterwards). Hence, autoconvolution is defined as
\begin{equation}
\mathbf{X} \ast \mathbf{X} = \mathcal{F}^{-1}(\mathcal{F}(\mathbf{X})^2),
\label{eq_autoconv_1}
\end{equation}
where $\mathcal{F}^{-1}$ - the $N$-dimensional inverse DFT, $\ast$ - convolution.
Because of squared frequencies, some phase information is lost and the inverse operation becomes ill-posed \citep{gorenflo1994autoconvolution}. 
In this work, we do not address this problem.

To extract patterns from $ \mathbf{X} $, it is necessary to make sure that $ mean(\mathbf{X})=0 $ and $ std(\mathbf{X}) > 0 $ before computing (\ref{eq_autoconv_1}). Also, to compute linear autoconvolution, $ \mathbf{X} $ is first zero-padded, i.e. for one dimensional case $ \mathbf{X} $ have to be padded with zeros to length $ (2s - 1)$, where $s$ - length of $ \mathbf{X} $ before padding.

\begin{figure}[t]
	\hspace{-10pt}
	\begin{minipage}[c][3.4cm][t]{0.5\textwidth}
		\centering
		\captionsetup[subfigure]{labelformat=empty, font=small,skip=-7pt}
		\subcaptionbox{\label{sub-mnist-filters}}{
			\captionsetup[subfigure]{labelformat=empty, font=footnotesize, position=top,skip=-1pt}
			\subcaptionbox{\hspace{8pt} $n=0$}{%
				\adjustbox{valign=m}{\raisebox{-5pt}{\small (a)}}
				\adjincludegraphics[valign=b,scale=\scaleFilters]{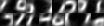}%
			}
			\subcaptionbox{$n=1$}{%
				\includegraphics[scale=\scaleFilters]{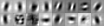}%
			}
			\subcaptionbox{$n=2$}{%
				\includegraphics[scale=\scaleFilters]{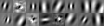}%
			}
			\subcaptionbox{$n=3$}{%
				\includegraphics[scale=\scaleFilters]{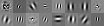}%
			}
		}
		\captionsetup[subfigure]{labelformat=empty,font=small,position=bottom,skip=-4pt}
		\subcaptionbox{\label{sub-mnist-filters-whitened}}{
			\captionsetup[subfigure]{labelformat=empty, font=footnotesize, position=bottom,skip=-7pt}
			\subcaptionbox{}{%
				\adjustbox{valign=m}{\raisebox{-10pt}{\small (b)}}
				\adjincludegraphics[valign=m,scale=\scaleFilters]{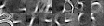}%
			}
			\subcaptionbox{}{%
				\includegraphics[scale=\scaleFilters]{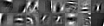}%
			}
			\subcaptionbox{}{%
				\includegraphics[scale=\scaleFilters]{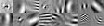}%
			}
			\subcaptionbox{}{%
				\includegraphics[scale=\scaleFilters]{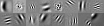}%
			}
		}
		\captionsetup[subfigure]{labelformat=empty, font=small,position=bottom,skip=-10pt}
		\subcaptionbox{\label{sub-cifar10-filters}}{
			\captionsetup[subfigure]{labelformat=empty, font=footnotesize, position=bottom,skip=-7pt}
			\subcaptionbox{}{%
				\adjustbox{valign=m}{\raisebox{-10pt}{\small (c)}}
				\adjincludegraphics[valign=m,scale=\scaleFilters]{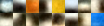}%
			}
			\subcaptionbox{}{%
				\includegraphics[scale=\scaleFilters]{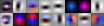}%
			}
			\subcaptionbox{}{%
				\includegraphics[scale=\scaleFilters]{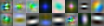}%
			}
			\subcaptionbox{}{%
				\includegraphics[scale=\scaleFilters]{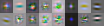}%
			}
		}
		\captionsetup[subfigure]{labelformat=empty,font=small,position=bottom,skip=0pt}
		\subcaptionbox{\label{sub-mnist-cifar10-whitened}}{
			\captionsetup[subfigure]{labelformat=empty, font=footnotesize, position=bottom,skip=-7pt}
			\subcaptionbox{}{%
				\adjustbox{valign=m}{\raisebox{-10pt}{\small (d)}}
				\adjincludegraphics[valign=m,scale=\scaleFilters]{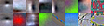}%
			}
			\subcaptionbox{}{%
				\includegraphics[scale=\scaleFilters]{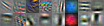}%
			}
			\subcaptionbox{}{%
				\includegraphics[scale=\scaleFilters]{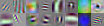}%
			}
			\subcaptionbox{}{%
				\includegraphics[scale=\scaleFilters]{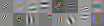}%
			}
		}
	\end{minipage}
	\vspace{-10pt}
	\caption{
		{Examples of filters $ \mathbf{D}^{(1)}$ of size $13 \times 13$ pixels learned using $k$-means on the MNIST (a,b) and CIFAR-10 (c,d) patches with (b,d) and without (a,c) whitening for recursive autoconvolution orders $n=0,1,2,3$. In each case, 16 filters are learned. Note that filters are more sparse for higher $n$. Similar filters are learned in the first layer of our AutoCNN networks.} 
	}
 	\vspace{-10pt}
	\label{fig_main2}
\end{figure}
 
\subsection{Recursive autoconvolution}

We adopt the recursive autoconvolution (RA) operator, proposed earlier in \citep{knyazev2014convolutional}, which is an extension of (\ref{eq_autoconv_1}):
\begin{equation}
\mathbf{X}_n = \mathbf{X}_{n-1} \ast \mathbf{X}_{n-1} = \mathcal{F}^{-1}(\mathcal{F}(\mathbf{X}_{n-1})^2),
\label{eq_rec_autoconv_2}
\end{equation}
where $ n=[0,n_{max}] $ - an index of the recursive iteration (or autoconvolution order). If $n=0$,  $\mathbf{X}_0$ equals the input, i.e. a raw image patch. For $n=1$ Eq. \ref{eq_rec_autoconv_2} becomes equal Eq. \ref{eq_autoconv_1}. In our work, we limit $n_{\max} = 3$ as higher orders do not lead to better classification results.

In \citep{knyazev2014convolutional}, image patterns extracted using this operator were used for parametric description of images. The \textit{novelty} of this work is that we use extracted patterns as convolution kernels, i.e. filters, because we noticed that applying Eq. \ref{eq_rec_autoconv_2} with $ n \ge 1 $ to image patches provides sparse wavelet-like patterns (Fig. \ref{fig_main}, \ref{fig_main2}), which are usually learned by a CNN in the first layer (see 
\citep{mairal2014convolutional}, Fig. 3 or \citep{wang2015unsupervised}, Fig. 6) or by other unsupervised learning methods, e.g., ICA (see \citep{le2011ica}, Fig. 1), or sparse coding (see \citep{labusch2008simple}, Fig. 2).
In addition, these patterns are similar to more complex mid-level features of CNNs (e.g., visualized in Fig. 2 in \citep{zeiler2014visualizing}).

\section{AutoCNN architecture}
\label{section_architecture}
\begin{figure}[!t]
	\centering
	\captionsetup[subfigure]{skip=1.5pt}
	{\includegraphics[scale=0.65, clip, trim=2cm 12.5cm 12.5cm 2.7cm]{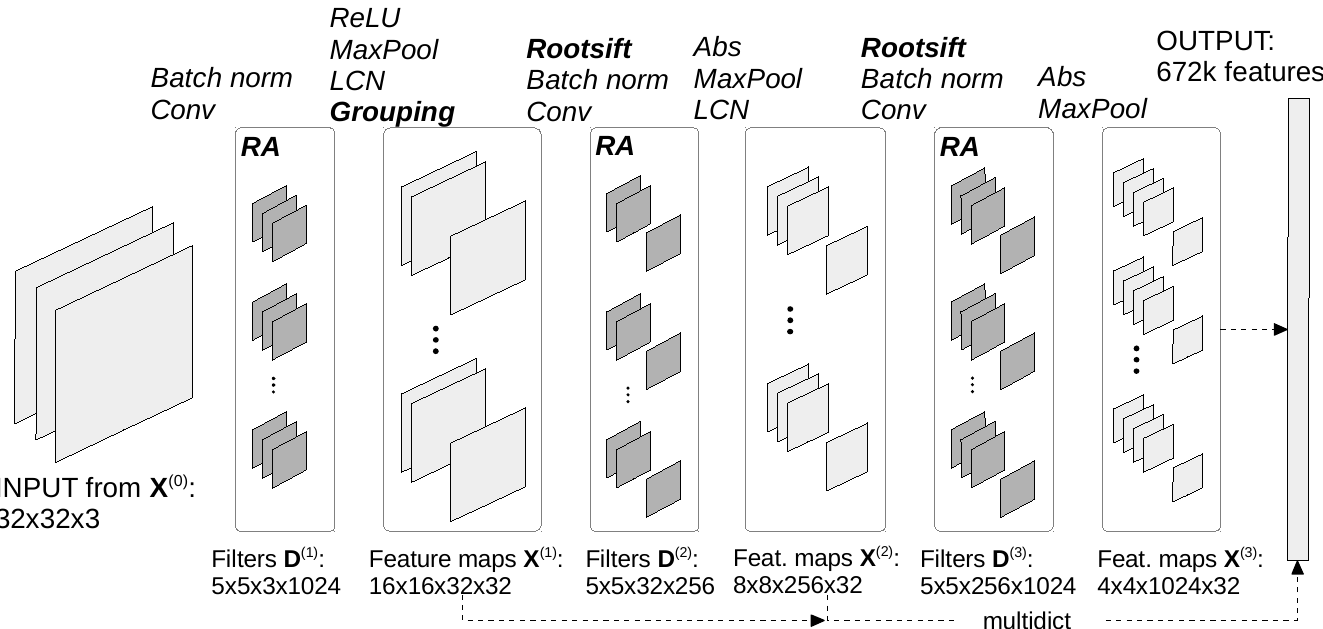}}
	\caption{Overview of our best architecture for CIFAR-10 (AutoCNN-L32). Filters $\mathbf{D}^{(1)}$, $\mathbf{D}^{(2)}$ and $\mathbf{D}^{(3)}$ are learned with unsupervised learning and recursive autoconvolution (RA). Features $\mathbf{X}^{(1)}$, $\mathbf{X}^{(2)}$ are pooled and concatenated with $\mathbf{X}^{(3)}$ to form a large feature vector (672k values). Steps that are usually missing in CNNs are \textbf{in bold}. }
	\label{fig:overview}
\end{figure}

In this section, we describe the architecture of a multilayer convolutional neural network (AutoCNN), which we train in an unsupervised way and then use as a feature extractor (Fig. \ref{fig:overview}). It is based on classical CNNs \citep{lecun1998gradient}, however, there is no backward pass and neither filters nor connections are trained using labels. In addition, based on \citep{dundar2015convolutional, miclut2014committees} we design two networks (AutoCNN-S32 and AutoCNN-L32) with features randomly split into several (32) groups (see Section \ref{section:grouping}), which is not typical for CNNs, but important in our work.
Since we have only forward pass, we are not limited by differentiable functions and can add nonlinearities such as Rootsift normalization ($\mathrm{sign}(\mathbf{x})\sqrt{| \mathbf{x} |/ \| \mathbf{x} \|_1 }$) \citep{arandjelovic2012three}, which improves classification significantly for some datasets. 
Batch standardization \footnote{Along this work, vector $\mathbf{x}$ is considered standardized if its $mean(\mathbf{x})=0$ and $std(\mathbf{x})=1$.} (or 'batch norm') is employed for all networks before convolutions - an entire batch is treated as a vector. In some cases (see details in Table \ref{table_dataset_details}) we follow \citep{miclut2014committees} and apply Local Contrast Normalization (LCN) between layers, but we do it after pooling for speedup.

The architectures of all networks used in this work are summarized in Table \ref{table_arch_details}.

\subsection{Training the network}
\label{section_learning}

Training our network is analogous to previous works on layer-wise learning \citep{dundar2015convolutional, miclut2014committees}.
We start by learning filters $\mathbf{D}^{(1)}$ for layer $l=1$ using training (unlabeled) images $\mathbf{X}^{(0)}$, then process these images with the learned filters $\mathbf{D}^{(1)}$ followed by rectification, pooling, normalization and further steps (depending on the architecture) that yield the first layer feature maps $\mathbf{X}^{(1)}$. Using these features, we train filters $\mathbf{D}^{(2)}$ for layer 2, and then process them $\mathbf{X}^{(1)}$ with the learned filters $\mathbf{D}^{(2)}$, and so forth. Features concatenated from all layers (multidictionary features) are used for classification with support vector machines (SVM).

Training filters $ \mathbf{D}^{(1)} \in \mathbb{R}^{s_1 \times s_1 \times d_1 \times K_1}$ for layer 1, that is $K_1$ filters of size $s_1 \times s_1$ with $d_1$ color channels, is performed in three steps: (1) random patches $\mathbf{X}^{(1)} \in \mathbb{R}^{s_1 \times s_1 \times d_{1} }$ are extracted from all training (unlabeled) samples; (2) recursive autoconvolution (Eq. \ref{eq_rec_autoconv_2}) is applied to them; (3) one of the unsupervised learning methods is applied to the autoconvolutional patches. Filters for the following layers are trained according to this procedure as well. 
Unless otherwise stated, $k$-means with Euclidean distance is employed for filter learning.

\subsubsection{Learning with recursive autoconvolution (RA)}
Learning filters with RA is a novel idea, so we explain some steps of its efficient application to our tasks.
One of the issues with RA is that the spatial size of image patches $\mathbf{X}$ is doubled after each iteration $n$ due to zero-padding (see Eq. \ref{eq_autoconv_1} and \ref{eq_rec_autoconv_2}). To learn filters from patches, we need all the patches to have some fixed size, so we simply take the central part of the result or resize (subsample) it to its original size after each iteration (Fig. \ref{fig_main}, where the second option is picked).
We randomly choose one of these options to make the set of patches richer.

\begin{figure*}[htp]
	\centering
	\hypersetup{
		colorlinks=true,
		linkcolor=black,
		citecolor=black,
		filecolor=black,
		urlcolor=black,
		linkbordercolor = {1 1 1},
		pdfborder = {0 0 0}
	}
	\input{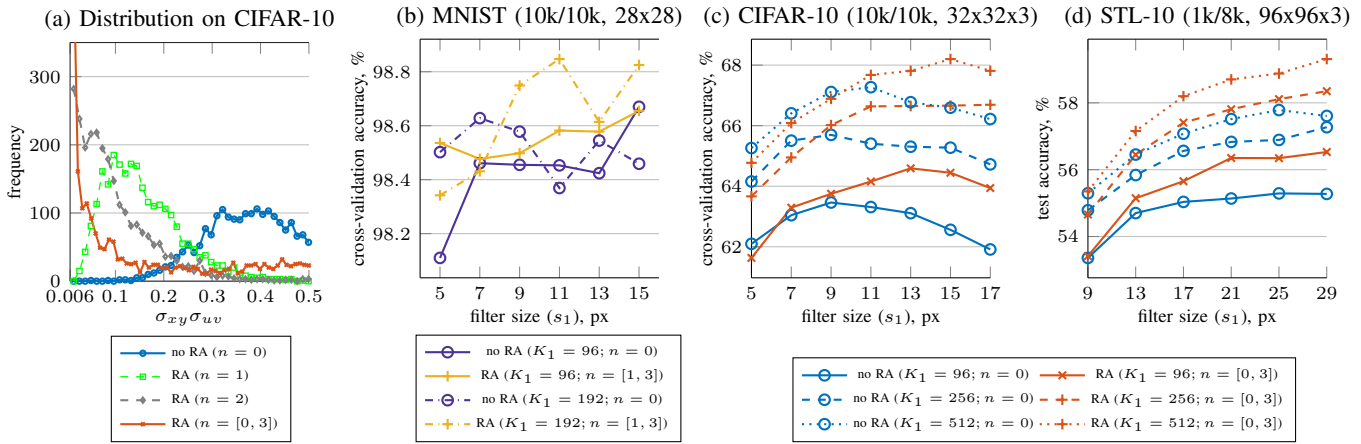} 
	\hspace{-0.1in} 
%
%
\definecolor{mycolor1}{rgb}{0.00000,0.44700,0.74100}%
\definecolor{mycolor2}{rgb}{0.65000,0.22500,0.04800}%
\definecolor{mycolor3}{rgb}{0.92900,0.69400,0.12500}%
\definecolor{mycolor4}{rgb}{0.3,0.2,0.6}%
\begin{tikzpicture}

\begin{axis}[%
width=1.25in,
height=1.25in,
at={(0.772in,0.516in)},
scale only axis,
legend columns=1,
legend entries={no RA ($K_1=96; n=0$), RA ($K_1=96; n=\lbrack \rangeConvOneThree \rbrack$), no RA ($K_1=192; n=0$), RA ($K_1=192; n=\lbrack \rangeConvOneThree \rbrack$)},
legend to name=namedRawMNIST,
ticklabel style = {font=\scriptsize},
xlabel={filter size ($s_1$), px},
ylabel={cross-validation accuracy, \%},
ylabel style = {font=\scriptsize, yshift=-5pt},
xlabel style = {font=\scriptsize, yshift=3pt},
legend style = {font=\tiny},
title={(b) MNIST (10k/10k, 28x28)},
ymajorgrids,
xtick={5, 7, 9, 11, 13, 15},
axis y line*=left,
title style={yshift=-5pt,font=\small},
]
\addplot [color=mycolor4,solid,line width=\lWidth,mark size=\mSize,mark=o,mark options={solid}]
table[row sep=crcr]{%
	5	98.11\\
	7	98.461\\
	9	98.455\\
	11	98.453\\
	13	98.424\\
	15	98.67\\
};
\addplot [color=mycolor3,solid,line width=\lWidth,mark size=\mSize,mark=+,mark options={solid}]
table[row sep=crcr]{%
	5	98.536\\
	7	98.477\\
	9	98.498\\
	11	98.582\\
	13	98.578\\
	15	98.655\\
};
\addplot [color=mycolor4,dashdotted,line width=\lWidth,mark size=\mSize,mark=o,mark options={solid}]
table[row sep=crcr]{%
	5	98.502\\
	7	98.628\\
	9	98.578\\
	11	98.369\\
	13	98.545\\
	15	98.459\\
};
\addplot [color=mycolor3,dashdotted,line width=\lWidth,mark size=\mSize,mark=+,mark options={solid}]
table[row sep=crcr]{%
	5	98.342\\
	7	98.431\\
	9	98.749\\
	11	98.847\\
	13	98.614\\
	15	98.825\\
};
\end{axis}
\end{tikzpicture}%
	\hspace{-0.1in} 
%
%
\definecolor{mycolor1}{rgb}{0.00000,0.44700,0.74100}%
\definecolor{mycolor2}{rgb}{0.85000,0.32500,0.09800}%
\definecolor{mycolor3}{rgb}{0.92900,0.69400,0.12500}%
\begin{tikzpicture}

\begin{axis}[%
width=1.25in,
height=1.25in,
at={(0.772in,0.516in)},
legend entries={no RA ($K_1=96; n=0$), RA ($K_1=96; n=\lbrack \rangeConvZeroThree \rbrack$), no RA ($K_1=256; n=0$), RA ($K_1=256; n=\lbrack \rangeConvZeroThree \rbrack$), no RA ($K_1=512; n=0$), RA ($K_1=512; n=\lbrack \rangeConvZeroThree \rbrack$)},
legend to name=namedRaw,
legend columns=2,
scale only axis,
ticklabel style = {font=\scriptsize},
xlabel={filter size ($s_1$), px},
ylabel={cross-validation accuracy, \%},
ylabel style = {font=\scriptsize, yshift=-5pt},
xlabel style = {font=\scriptsize, yshift=3pt},
legend style = {font=\tiny},
title={(c) CIFAR-10 (10k/10k, 32x32x3)},
ymajorgrids,
xtick={5, 7, 9, 11, 13, 15, 17},
xmin=5,
xmax=17,
axis y line*=left,
title style={yshift=-5pt,font=\small}
]
\addplot [color=mycolor1,solid,line width=\lWidth,mark size=\mSize,mark=o]
  table[row sep=crcr]{%
5	62.097\\
7	63.043\\
9	63.462\\
11	63.312\\
13	63.11\\
15	62.56\\
17	61.912\\
};
\addplot [color=mycolor2,solid,line width=\lWidth,mark size=\mSize,mark=x]
  table[row sep=crcr]{%
5	61.632\\
7	63.292\\
9	63.75\\
11	64.156\\
13	64.591\\
15	64.445\\
17	63.94\\
};
\addplot [color=mycolor1,dashed,line width=\lWidth,mark size=\mSize,mark=o,mark options={solid}]
table[row sep=crcr]{%
	5	64.156\\
	7	65.506\\
	9	65.695\\
	11	65.405\\
	13	65.305\\
	15	65.273\\
	17	64.722\\
};
\addplot [color=mycolor2,dashed,line width=\lWidth,mark size=\mSize,mark=+,mark options={solid}]
table[row sep=crcr]{%
	5	63.665\\
	7	64.951\\
	9	66.019\\
	11	66.638\\
	13	66.649\\
	15	66.662\\
	17	66.691\\
};
\addplot [color=mycolor1,dotted,line width=\lWidth,mark size=\mSize,mark=o,mark options={solid}]
table[row sep=crcr]{%
	5	65.261\\
	7	66.407\\
	9	67.112\\
	11	67.271\\
	13	66.775\\
	15	66.593\\
	17	66.219\\
};
\addplot [color=mycolor2,dotted,line width=\lWidth,mark size=\mSize,mark=+,mark options={solid}]
table[row sep=crcr]{%
	5	64.773\\
	7	66.086\\
	9	66.886\\
	11	67.676\\
	13	67.81\\
	15	68.2\\
	17	67.811\\
};
\end{axis}
\end{tikzpicture}%
	\hspace{-0.2in}
%
%
\definecolor{mycolor1}{rgb}{0.00000,0.44700,0.74100}%
\definecolor{mycolor2}{rgb}{0.85000,0.32500,0.09800}%
\definecolor{mycolor3}{rgb}{0.92900,0.69400,0.12500}%
\begin{tikzpicture}

\begin{axis}[%
width=1.25in,
height=1.25in,
at={(0.772in,0.516in)},
scale only axis,
ticklabel style = {font=\scriptsize},
xlabel={filter size ($s_1$), px},
ylabel={test accuracy, \%},
ylabel style = {font=\scriptsize, yshift=-5pt},
xlabel style = {font=\scriptsize, yshift=3pt},
title={(d) STL-10 (1k/8k, 96x96x3)},
ymajorgrids,
xtick={9, 13, 17, 21, 25, 29},
xmin=9,
xmax=29,
axis y line*=left,
title style={yshift=-5pt,font=\small},
legend style={anchor=north east, font=\tiny},
legend pos=outer north east
]
\addplot [color=mycolor1,solid,line width=\lWidth,mark size=\mSize,mark=o]
table[row sep=crcr]{%
9	53.36\\
13	54.70125\\
17	55.03625\\
21	55.1375\\
25	55.29\\
29	55.275\\
};
\addplot [color=mycolor2,solid,line width=\lWidth,mark size=\mSize,mark=x]
table[row sep=crcr]{%
9	53.415\\
13	55.14625\\
17	55.655\\
21	56.35\\
25	56.345\\
29	56.52875\\
};
\addplot [color=mycolor1,dashed,line width=\lWidth,mark size=\mSize,mark=o,mark options={solid}]
table[row sep=crcr]{%
	9	54.78625\\
	13	55.83875\\
	17	56.5625\\
	21	56.83125\\
	25	56.89\\
	29	57.27125\\
};
\addplot [color=mycolor2,dashed,line width=\lWidth,mark size=\mSize,mark=x,mark options={solid}]
table[row sep=crcr]{%
	9	54.66\\
	13	56.4475\\
	17	57.4075\\
	21	57.80875\\
	25	58.11125\\
	29	58.34875\\
};
\addplot [color=mycolor1,dotted,line width=\lWidth,mark size=\mSize,mark=o,mark options={solid}]
table[row sep=crcr]{%
	9	55.295\\
	13	56.4525\\
	17	57.07625\\
	21	57.51375\\
	25	57.78375\\
	29	57.6075\\
};
\addplot [color=mycolor2,dotted,line width=\lWidth,mark size=\mSize,mark=+,mark options={solid}]
table[row sep=crcr]{%
	9	55.35125\\
	13	57.1575\\
	17	58.19875\\
	21	58.70125\\
	25	58.87625\\
	29	59.31\\
};
\end{axis}
\end{tikzpicture}%
	\hspace*{15pt}
	\ref{namedRawSigma}
	\hspace{40pt}
	\ref{namedRawMNIST}
	\hspace{35pt}
	\ref{namedRaw}
	\caption{(a) Recursive autoconvolution (RA) makes filters more localized and sparse as shown by distributions of the joint spatial and frequency resolution $\sigma_{xy}\sigma_{uv}$ for the first layer filters learned on CIFAR-10 with $k$-means with different orders $n$.
		(b-d) For our simple single layer network (AutoCNN-S1) RA improves results for a wide range of filter sizes $s_1$ and number of filters ($K_1$) on three datasets: MNIST (b), CIFAR-10 (c), STL-10 (d). Not that in some cases results with RA are better than without RA even if the latter has two times more filters. The number of training/test samples and input image sizes in the datasets are indicated in parentheses for reference.}
	\label{fig_acc_vs_fiter_size_raw}
\end{figure*}

Next, according to our statistics of extracted patches, presented in Fig. \ref{fig_acc_vs_fiter_size_raw}(a),
autoconvolution order $n$ is inversely proportional to the joint spatial $\sigma_{xy}$ and frequency $\sigma_{uv}$ resolution, i.e. $ n \sim 1/(\sigma_{xy}\sigma_{uv}),$
%
%
where $\sigma_{xy}=\sigma_{x}\sigma_{y}=\sqrt{D_1D_2}$ and $D_1,D_2$ - are eigenvalues of the weighted covariance matrix of $\mathbf{X}$ in the spatial domain; analogously for $\sigma_{uv}$. Therefore, to cover a wider range of spatio-frequency properties and to learn a more diverse set of filters, patches extracted with several orders are combined into one global set. That is, we take results of several orders (e.g., in case $n=[0,3]$ we have 4 patches instead of 1) and put them into one global set of autoconvolutional patches.
Note that in case $n=0$, we extract more patches to make the total number of input data points for $k$-means about the same as for combinations of orders. 
In this global set, all patches are first scaled to have values in the range [0,1], then they are ZCA-whitened 
as in \citep{coates2011analysis, dundar2015convolutional, miclut2014committees, krizhevsky2009learning}.
Even though, with RA we can extract Gabor-like patterns without whitening (see Fig. \ref{fig_main2}(a),(c)), such preprocessing is essential for $k$-means to learn less correlated filters.
For this whitened set $k$-means produces a set of $K_l$ data points (a dictionary) $ \mathbf{D}^{(l)} \in \mathbb{R}^{s_l \times s_l \times d_l \times K_l}$.
These data points are first $l_2$-normalized and then used as convolution kernels (filters) for layer $l$. 

\subsubsection{Grouping of feature maps}
\label{section:grouping}
Filters of the second and following layers can be trained either in the same way as of the first layer or using a slightly modified procedure borrowed from \citep{dundar2015convolutional, miclut2014committees}.
Concretely, features $\mathbf{X}^{(l)}, l>1$ can be split (randomly in this work) into $G_l$ groups. It is useful in practice, because filters of layer $l+1$ will have smaller depth $d_{l+1}=K_{l}/G_l$ (instead of $d_{l+1}=K_{l}$ as in typical CNNs) and smaller overall dimensionality, so that it is easier to learn such filters with $k$-means. At the same time, the number of features becomes larger by factor $G_l$, and more features usually improve classification in case of unsupervised learning. 

\subsubsection{Shared filters for $G_l > 1$}
\label{section:shared_filters}
The disadvantage of such splitting is that filters have to be learned for each group independently, i.e. it is necessary to run $k$-means $G_l$ times.
For instance, for the second layer we have $ \mathbf{D}^{(2)} \in \mathbb{R}^{s_2 \times s_2 \times d_2 \times K_2 \times G_1}$, i.e. the total number of filters equals $K_2 \times G_1$. Our contribution is that instead of treating groups independently, we learn filters for all groups taken together without negative effect on classification accuracy (see results in Table \ref{table:shared}).
In this case, patches from all $G_1$ feature map groups are concatenated before clustering and $k$-means is run only once. Thus, the total number of filters in layer 2 becomes equal $K_2$, i.e. the same filters are \textit{shared} between all $G_1$ groups. This trick enables us to learn our large AutoCNN-L32 with $G_1=32$ in a feasible time.

\section{Experiments}
\label{section:experiments}

\subsection{Experimental setup}
We evaluate our method on four image classification benchmarks: MNIST \citep{lecun1998gradient}, CIFAR-10, CIFAR-100 \citep{krizhevsky2009learning} and STL-10 \citep{coates2011analysis}. 
To demonstrate that unsupervised learning is particularly effective for datasets with few training samples, such as STL-10, which has only 100 training images per class, we test our method on reduced versions of MNIST and CIFAR-10, namely MNIST (100), MNIST (300) and CIFAR-10 (400) with just 100, 300 and 400 training images per class respectively. We follow the same experimental protocol as in previous works, e.g., \citep{dundar2015convolutional, dosovitskiy2014discriminativeLatest}: 10 random subsets (folds) from the training set are drawn, while the test set remains fixed. For STL-10 these folds are predefined.
Unless otherwise stated, we report average classification accuracies (or errors on MNIST) in percent; on reduced datasets, these results are averaged over 10 runs. In Tables \ref{table:learning_methods}-\ref{table_test_results_cifar_stl10} better results are indicated in bold.
Images of all datasets, except for MNIST, are ZCA-whitened as in most previous works (see details in Table \ref{table_dataset_details}).

While in previous works all labeled training samples are typically used as unlabeled data during unsupervised learning, we found that it is enough to use at most 10k-20k samples to learn filters in all experiments, including STL-10, which contains 100k unlabeled samples.


We train models with 1-3 layers according to the proposed architecture (Section \ref{section_architecture}). The details of all models are presented in Table \ref{table_arch_details}. Network parameters (number of layers and filters, pooling size and stride, etc.) are chosen so that to make them more consistent with previous works
\citep{dosovitskiy2014discriminativeLatest,makhzani2015winner,miclut2014committees,dundar2015convolutional} and within this work.

In case of a multilayer network, we use multidictionary features for classification, which is a standard approach in unsupervised learning \citep{dundar2015convolutional,lin2014stable,dosovitskiy2014discriminativeLatest}. For example, in case of three layers, features of bottom layers ($l=1,2$) are pooled so that their spatial size equals the size of the third layer features, afterwards they are concatenated.  

In all cases, except for AutoCNN-L32, a linear SVM is used as a classifier.

\subsection{Effect of filter size and number of filters} 

First, we experiment with a simple single layer architecture (AutoCNN-S1) to see the effect of recursive autoconvolution (Eq. \ref{eq_rec_autoconv_2}) on classification performance depending on the filter size ($s_1$) and the number of filters ($K_1$) (Fig. \ref{fig_acc_vs_fiter_size_raw} (b)-(d)). In this experiment, in case of MNIST and CIFAR-10, 10-fold cross-validation is performed with 10k training and 10k test samples drawn from the original training sets. For STL-10 we use the original predefined folds.
We observe that recursive autoconvolution consistently improves classification. It is evident especially for larger filters, because RA makes filters more localized and sparse (Fig. \ref{fig_acc_vs_fiter_size_raw}(a)). This experiment also shows that the results generalize well across different datasets and preprocessing strategies.

We also experimented with each of the orders in Eq. \ref{eq_rec_autoconv_2} independently (results not shown) to determine filters of which orders contribute the most during classification, but we found that, generally, combinations of orders work best.

\begin{table}[htbp]
	\renewcommand{\arraystretch}{0.8}
	\centering
	\caption{(a) Comparison of learning algorithms on CIFAR-10(400) using AutoCNN-S1-256.
		(b) Shared vs independent filters on CIFAR-10(400) for a two layer model AutoCNN-S32 + RA ($N$ - number of trainable parameters, 
		\textsuperscript{†}: time for learning filters with $k$-means for 32 groups measured on Intel Xeon CPU E5-2620v3.). }
	\footnotesize
	\centering
	\captionsetup[subtable]{skip=0pt}
	\begin{subtable}{.5\linewidth}
	{\setlength{\tabcolsep}{3.0pt}
	\caption{}
	\label{table:learning_methods}
	\begin{tabular}{lcc}
		\toprule
		Learning method 
		& Raw & RA \\
		\midrule
		$k$-means & 64.0 & \bf 65.6 \\
		PCA ($s_1 \ge 11$) & \bf 60.7 & 59.8 \\
		ICA ($s_1 \ge 11$) & 62.8 & \bf 64.5 \\
		\bottomrule
	\end{tabular}}
	\end{subtable}
	\begin{subtable}{.45\linewidth}
	{\setlength{\tabcolsep}{-1.0pt}
	\caption{}
	\label{table:shared}
	\begin{tabular}{lccc}
		\toprule
		Filters & Avg acc & $N$ & Train $t$ \textsuperscript{†} \\
		\midrule
		Independ. & \bf 72.0 & 432.5k & 470s \\
		Shared & 71.9 & 43.6k & 20s \\
		\bottomrule
	\end{tabular}}
	\end{subtable}
	\hspace{-5pt}
\end{table}

\subsubsection{Recursive autoconvolution and ICA or PCA}
To demonstrate generalizability of our method, we investigated if recursive autoconvolution is able to improve other learning methods. For this purpose, we trained $K_1=256$ filters with principal (PCA) and independent (ICA) component analysis 
\citep{hyvarinen1999fast} on patches with $n=0$ (Raw) and $n=[0,3]$ (RA) using the same procedure as with $k$-means (Section \ref{section_learning}). Filter sizes have to be chosen based on results from the previous experiment (Fig. \ref{fig_acc_vs_fiter_size_raw} (b)-(d)).
However, we had to use larger filters to satisfy $s_1 \times s_1 \times 3 \geq K_1$, since we did not learn overcomplete dictionaries. Even though RA does not improve PCA filters for some reasons, in case of ICA the gap between results with and without RA is notable (Table \ref{table:learning_methods}), which confirms the effectiveness of our method.

\subsection{Multilayer performance}
\label{section_multilayer_test}

We performed a series of experiments with convolutional architectures that match or similar to the ones in the previous works on unsupervised learning \citep{dosovitskiy2014discriminativeLatest,makhzani2015winner}. It allows us to fairly compare features and to see the benefit of RA for a multilayer AutoCNN.

\subsubsection{Comparison to Exemplar-CNN}

Compared to Exemplar-CNNs \citep{dosovitskiy2014discriminativeLatest}, our results appear to be slightly inferior at first (Table \ref{table:exemplar}), but our networks have only 3 layers and we do not exploit data augmentation so extensively. For instance, we do not use any color augmentation (see details in Table \ref{table_dataset_details}). To improve our results, we trained a larger 3 layer network (AutoCNN-L), and were able to outperform Exemplar-CNNs on CIFAR-10(400). 

\newcolumntype{R}[1]{>{\Centering\hspace{-8pt}}p{#1}<{\hspace{-12pt}}}

\begin{table}[htbp]
	\renewcommand{\arraystretch}{0.8}
	\small
	\centering
	\caption{Comparison of our features with and without RA to Exemplar-CNN \citep{dosovitskiy2014discriminativeLatest} (A: data augmentation, A$^{*}$: large data augmentation, average accuracy (\%) for 10 folds is reported).}
	\label{table:exemplar}
	\footnotesize\setlength{\tabcolsep}{3.0pt}
	\centering
	\begin{tabular}{l@{\hspace{5pt}} *{5}{c}}
		\toprule
		Model 
		& \multicolumn{2}{c}{CIFAR-10(400)} & \multicolumn{2}{c}{STL-10} & \# filters  \\
		\midrule
		Exemplar-CNN + A$^{*}$
		&  \multicolumn{2}{c}{75.7} & \multicolumn{2}{c}{74.9} & 64-128-256-512 \\ 
		Exemplar-CNN + A$^{*}$
		& \multicolumn{2}{c}{77.4}  & \multicolumn{2}{c}{75.4} & 92-256-512-1024\\
		\midrule
		& Raw & RA & Raw & RA & \\
		\midrule
		AutoCNN-S & 66.5 & \bf 68.4 & 60.6 & \bf 62.3 & 64-128-256 \\
		AutoCNN-S - Rootsift & 60.9 & \bf 61.0 & 51.9 & \bf 52.5 & 64-128-256 \\
		AutoCNN-S + A & 69.9 & \bf 72.1 & 67.6 & \bf 69.2 & 64-128-256\\
		AutoCNN-M + A & 72.2 & \bf 74.5 & $69.7$ & \bf 70.6 & 92-256-512 \\
		AutoCNN-L + A & 75.9 & \bf 77.6 & 72.4 & \bf 73.1 & 256-1024-2048\\
		\bottomrule
	\end{tabular}
\end{table}

Note, however, that for all models in Table \ref{table:exemplar} the results with RA are better than without (columns 'Raw'), while all model settings are kept identical. 
In fact, for CIFAR-10(400) the network without recursive autoconvolution has to be virtually doubled in size in order to catch up wih the score of the network with RA (compare results of AutoCNN-S and AutoCNN-M).

Using a small AutoCNN-S network we estimated the contribution of Rootsift \citep{arandjelovic2012three} to our results by removing it from our pipeline. Not only it boosts classification, but also allows recursive autoconvolution to work more efficiently for multilayer models. 
However, note that our plots in Fig. \ref{fig_acc_vs_fiter_size_raw} and all results on MNIST are obtained without it, so Rootsift is not always required.

\subsubsection{Comparison to CONV-WTA} Compared to the Winner Takes All Autoencoder (CONV-WTA) \citep{makhzani2015winner}, our networks are more efficient even with a smaller number of filters (Table \ref{table:conv_wta}). With RA our results are further improved. In a few cases we had to project features with PCA to 4096 (PCA4k) dimensions before classification to fit all CIFAR-10 features into memory. It is an undesirable operation, because PCA degrades our features.

We also compare our models to CONV-WTA on MNIST(100) and full MNIST (Table \ref{table_test_results_mnist}). Two layer AutoCNN-S2 with recursive autoconvolution yields better results than CONV-WTA, even though our model has two times fewer filters in the second layer (other parameters are the same, Rootsift is not used on MNIST). Our error of 0.39\% on full MNIST is competitive even when compared to supervised learning. For instance, the same error was reported in \citep{mairal2014convolutional} and 0.47\% in \citep{zeiler2013stochastic} with 2 and 3 layer convolutional networks respectively. Without RA our model performs significantly worse.

\begin{table}[htbp]
	\renewcommand{\arraystretch}{0.8}
	\centering
	\caption{Comparison of our features with and without RA to CONV-WTA \citep{makhzani2015winner} on full CIFAR-10 (A: data augmentation).}
	\label{table:conv_wta}
	\footnotesize
	\setlength{\tabcolsep}{2.0pt}
	\centering
	\begin{tabular}{l@{\hspace{5pt}} *{3}{c}}
		\toprule
		Model & \multicolumn{2}{c}{CIFAR-10} & \# filters  \\
		\midrule
		CONV-WTA & \multicolumn{2}{c}{72.3} & 256 \\
		CONV-WTA & \multicolumn{2}{c}{77.9} & 256-1024 \\
		CONV-WTA & \multicolumn{2}{c}{80.1} & 256-1024-4096 \\
		\midrule
		& Raw & RA & \\
		\midrule
		AutoCNN-S1-256 & 71.3 & \bf 72.4 & 256 \\
		AutoCNN-M2 & 80.0 & \bf 80.4 & 256-1024 \\
		AutoCNN-M2 + PCA4k & 77.4 & \bf 78.9 & 256-1024 \\
		AutoCNN-L + PCA4k & 80.1 & \bf 81.4 & 256-1024-2048\\
		AutoCNN-L + PCA4k + A & 82.7 & \bf 84.4 & 256-1024-2048 \\
		\bottomrule
	\end{tabular}
\end{table}

\begin{table}[htbp]
	\renewcommand{\arraystretch}{0.8}
	\centering
	\caption{Classification errors on MNIST using unsupervised methods (for MNIST(100) the format is  $error \pm std$).}
	\label{table_test_results_mnist}
	\footnotesize\setlength{\tabcolsep}{5.0pt}
	\begin{tabular}{l@{\hspace{-3pt}} *{3}{c}}
		\toprule
		Model & MNIST (100) & MNIST & \# filters \\
		\midrule
		Sparse coding \citep{labusch2008simple}  & $-$ & 0.59 & 169\\
		C-SVDDNet \citep{wang2014unsupervised} 
		& $-$ & 0.35 & 400 \\
		CONV-WTA \citep{makhzani2015winner} & $-$ & 0.64 & 128\\
		CONV-WTA \citep{makhzani2015winner} & 1.92 & 0.48 & 128-2048 \\
		\midrule
		AutoCNN-S1-128 & 2.89 $\pm$ 0.16  & 0.82 & 128 \\
		AutoCNN-S1-128 + RA & 2.45 $\pm$ 0.10 & 0.69 & 128 \\
		AutoCNN-S2 & 1.91 $\pm$ 0.10 &  0.57 & 128-1024 \\
		AutoCNN-S2 + RA & \bf 1.75 $\pm$ 0.10 & \bf 0.39 & 128-1024\\
		\midrule
		\midrule
		Semi-supervised best & $ 0.84 \pm 0.08$ \citep{rasmus2015semi} & 0.5 \citep{dundar2015convolutional} & \\
		\bottomrule
	\end{tabular}
\end{table}

\begin{table*}[t!]
	\renewcommand{\arraystretch}{0.8}
	\caption{Classification accuracies on the test sets using unsupervised methods (A: data augmentation, \textsuperscript{†}: obtained with a larger network and supervised learning of connections between layers).}
	\label{table_test_results_cifar_stl10}
	\centering
	\footnotesize
	\setlength{\tabcolsep}{5.0pt}
	\begin{tabular}{l@{\hspace{3pt}} *{5}{c}}
		\toprule
		Model & CIFAR-10(400) & CIFAR-10 & STL-10 & CIFAR-100 & \# filters \\
		\midrule
		NOMP-20 \citep{lin2014stable} & $72.2 \pm 0.4 $ & 82.9 & $67.9 \pm 0.6$ & $60.8$ & 3200-6400-6400 \\
		Conv. clustering, unsup. 
		\citep{dundar2015convolutional} & $-$ & $-$ & $65.4$ ($74.1$\textsuperscript{†}) & $-$ & 96-1536 \\
		CONV-WTA \citep{makhzani2015winner} & $-$ & $80.1$ & $-$ & $-$ & 256-1024-4096 \\
		Committee of nets + A \citep{miclut2014committees} & $-$ & $-$ & $68.0 \pm 0.6$ & $-$ & 300-5625 \\
		$k$-means + A \citep{hui2013direct} & $72.6 \pm 0.7 $ & 81.9 & $63.7$ & $-$ & 3 layers, 6400 \\
		Exemplar-CNN + A$^{*}$ \citep{dosovitskiy2014discriminativeLatest}  & $ 77.4 \pm 0.2$ & $84.3$ & $\bf 75.4 \pm 0.3 $  & $-$ & 92-256-512-1024 \\
		\midrule
		AutoCNN-L32 + RA + PCA1k & $76.4 \pm 0.4$ & $85.4$ & $68.7 \pm 0.5$ & $63.9$ & \multirow{3}{*}{1024-256-1024} \\
		AutoCNN-L32 + A + PCA1.5k & $78.2 \pm 0.3$ & $87.1$ & $74.0 \pm 0.6$ & $67.1$ & \\
		AutoCNN-L32 + RA + A + PCA1.5k & $\bf 79.4\pm0.3$ & $\bf 87.9$ &  $74.5 \pm 0.6$  & $\bf 67.8$ & \\
		\midrule
		\midrule
		Pretrained on ImageNet & $73.8 \pm 0.4$ \citep{radford2015unsupervised} & $89.1$ \citep{hertel2015deep} & $-$ & $67.7$ \citep{hertel2015deep} & \\
		Semi-supervised state of the art & $79.6\pm 0.5$ \citep{rasmus2015semi} & $ 92.2$ \citep{zhao2015stacked} & $74.3$ \citep{zhao2015stacked} & $69.1$ \citep{zhao2015stacked} & \\
		\bottomrule
	\end{tabular}
\end{table*}

\subsection{Performance of large AutoCNN-L32}
\label{section:large_exper}
Unsupervised layer-wise learning does not suffer from overfitting, but making layers wider than in our AutoCNN-L is computationally demanding. Meanwhile, splitting features into $G_l$ groups (see Section \ref{section:grouping}) provides more features, and in certain settings it also benefits classification. So we designed a large AutoCNN-L32 network with $G_l=32$ groups. Due to feature splitting and the idea of shared filters introduced earlier in this work (Section \ref{section:shared_filters} and Table \ref{table:shared}), training time remains comparable to AutoCNN-L. We also tried other architectures, but present results only with the best one.  

\subsubsection{Dimension reduction and classification}
\label{section_committee}
Output features of AutoCNN-L32 are prohibitively large, so we first apply randomized principal component analysis \citep{halko2011finding} together with whitening and then train an SVM on the projected data. In this case, we use an RBF-SVM in an efficient GPU implementation \citep{cotter2011gpu}. A nonlinear SVM is chosen, because it performs better with the projected features.
The SVM regularization constant is fixed to $C=16$ and the width of the RBF kernel is chosen to be $\gamma = 1/N_{PCA}$, where $N_{PCA}$ is the dimensionality of the projected feature vector, which is the input of the SVM.

\subsubsection{Evaluation}
AutoCNN-L32 achieves very competitive results on both reduced and full datasets (Table \ref{table_test_results_cifar_stl10}).
We found that for layers 2-3 of this particular model it is better to train filters  with PCA rather than with $k$-means. It is surprising, because according to our results with a single layer model (Table \ref{table:learning_methods}), PCA should be worse than $k$-means in this task.
But we obtain just $77.4 \pm 0.3$\% on CIFAR-10(400) with AutoCNN-L32 if $k$-means is used for all layers.
To obtain results on CIFAR-100 we use the same settings as on CIFAR-10.

\subsection{Analysis of results}	
Notably, in most of the previous works (see Tables \ref{table_test_results_mnist} and \ref{table_test_results_cifar_stl10}), the classification results are very good either for simple grayscale images (MNIST) or more complex colored datasets (CIFAR-10, CIFAR-100, STL-10) or for smaller or larger datasets only. 
The only exception seems to be Ladder Networks \citep{rasmus2015semi}, which are much larger and deeper than our models and are semi-supervised. Also, in \citep{wang2014unsupervised} better results are achieved on MNIST, but it can be quite easy to fine tune to such a simple task \footnote{See https://github.com/bknyaz/gabors}.
Our models are especially effective for CIFAR-10, showing performance close to Ladder Networks. 
Our results on this dataset (both reduced and full) are several percent higher (absolute difference) compared to other unsupervised models. On full CIFAR-10 and CIFAR-100 with 5000 and 500 training labels per class respectively, our models also outperform some advanced fully supervised CNNs \citep{zeiler2013stochastic, mairal2014convolutional} and better or comparable to CNNs pretrained in an unsupervised way \citep{paine2014analysis} or on the large-scale dataset with over 1 million of training samples \citep{radford2015unsupervised,hertel2015deep}. On STL-10 our network with RA compares favorably to the 10 layer convolutional autoencoder (SWWAE) \citep{zhao2015stacked}.
Other previous works, showing higher accuracies, are either fully or semi-supervised.

Our deeper networks outperform our shallow ones, which suggests the importance of depth in our case in the same way as in supervised CNNs. 

One of the drawbacks in our work is that our best results on CIFAR and STL-10 are obtained with a nonlinear SVM, but we believe that given the overall simplicity of our models it can be reasonable to use a more complex kernel. Moreover, in our experience the RBF kernel does not always improve classification results if used on top of CNN features, but does in our case. In addition, we show competitive results with a linear SVM in Tables \ref{table:learning_methods}-\ref{table_test_results_mnist}. We also do not thoroughly validate all model components (which would be out of scope), because the main intention of this work is to show that recursive autoconvolution boosts classification in a number of different settings and is important to achieve our best results.

In this work, we show good results both for simple and more complex datasets, as well as both for smaller and larger ones using the same model with few tuning parameters. Among unsupervised learning methods without data augmentation we report state of the art results for all datasets.
One of the main reasons for such results is that recursive autoconvolution allows us to learn a set of filters with rich spatio-frequency properties. In AutoCNNs, most of the learned filters have joint spatial and frequency resolution close to the theoretical minimum (Fig. \ref{fig_acc_vs_fiter_size_raw}(a), RA with $n=[0,3]$), i.e. they resemble simple edge detectors, such as Gabor filters. Other filters are uniformly distributed across a wide range of resolutions and typically have complex meaningful shapes (Fig. \ref{fig_main}, \ref{fig_main2}). By learning large sets of such filters our models are able to detect a lot of highly diverse features, so that an SVM can easily choose the discriminative ones.

\paragraph{Computational complexity.}
Note that the numbers of network filters presented in Tables \ref{table:exemplar}-\ref{table_test_results_cifar_stl10} do not always reflect the total number of trainable parameters ($N$) nor the total computational cost. While our 3 layer AutoCNN-L32 model has more filters than some CNNs, for convolutional layers it has
$N \approx 6.8\cdot10^6$ parameters, while both the largest Exemplar-CNN \citep{dosovitskiy2014discriminativeLatest} and CONV-WTA \citep{makhzani2015winner} models have $\approx 40\cdot10^6$ parameters (in case of CIFAR-10).
Training our large 3 layer network on full CIFAR-10 with data augmentation takes about 85 minutes on NVIDIA GTX 980 Ti, Intel Xeon CPU E5-2620v3 and 64GB RAM in a Matlab implementation with VLFeat \citep{vedaldi2010vlfeat}, MatConvNet \citep{vedaldi2015matconvnet} and GTSVM \citep{cotter2011gpu}. It's also important that computational cost of recursive autoconvolution for filter learning is less than 10\% of overall training time. 

Source codes to reproduce our results are available at \texttt{https://github.com/bknyaz/autocnn\_unsup}.

\section{Conclusion}
The importance of unsupervised learning in visual tasks is increasing and development is driven by the necessity
to better exploit massive amounts of unlabeled data.
We propose a novel idea for unsupervised feature learning and report competitive results in several image classification tasks among the works not relying on supervised learning. Moreover, we report state of the art results among unsupervised methods without data augmentation.
We adopt recursive autoconvolution and demonstrate its great utility for unsupervised learning methods, such as $k$-means and ICA. 
We argue that it can also be integrated into more recent learning methods, such as convolutional clustering, to boost their performance. Furthermore, we significantly reduce the total number of trainable parameters by using shared filters. As a result, the proposed autoconvolutional network performs better than most of the unsupervised, and several supervised, models in various classification tasks with only few, but also with thousands of, labeled samples.

\section*{Acknowledgment}
This work is jointly supported by the German Academic Exchange Service (DAAD) and the Ministry of Education and Science of the Russian Federation (project number 3708).

\medskip
\small
\bibliographystyle{unsrt}
\bibliography{./ijcnn17.bib}
\vfill

\begin{table*}[htpb]
	\renewcommand{\arraystretch}{1.0}
	\caption{Dataset and some model parameters used in the experiments (LCN - Local Contrast Normalization)}
	\label{table_dataset_details}
	\centering
	\scriptsize
	\begin{tabular}{ccccc}
		\toprule
		Dataset & Preprocessing & LCN (>1 layers) & Multidictionary (>1 layers) & Data augmentation (A) \\
		\midrule
		MNIST & $-$ & $-$ & $+$ & $-$ \\ 
		CIFAR-10, CIFAR-100 & ZCA-whitening & $+$ & $+$ & mirroring \\
		STL-10 & ZCA-whitening & $+$ & $+$ & mirroring, crops(72px), scaling(1-1.3), rotation($\pm$10 degrees) \\ 
		\bottomrule
	\end{tabular}
\end{table*}

\begin{table*}[ht!]
	\renewcommand{\arraystretch}{0.7}
	\caption{Network architectures used in the experiments}
	\label{table_arch_details}
	\centering
	\footnotesize\setlength{\tabcolsep}{2.0pt}
	\begin{tabular}{l@{\hspace{6pt}} *{20}{c}}
		\toprule
		Model & $L$ & $K_1$ & $n_1$ & $s_1,d_1$ & $R_1$ & $m_1$ & $G_1$ & $K_2$ & $n_2$ & $s_2,d_2$ & $R_2$ & $m_2$ & $G_2$ & $K_3$ & $n_3$ & $s_3,d_3$ & $R_3$ & $m_3$ & root & SVM \\
		\midrule
		\multicolumn{21}{c}{\bf MNIST} \\
		\midrule
		AutoCNN-S1 & 1 & $K_1$ & 1-3 & $s_1$x$s_1$x1 & $|x|$ & 4(4) & $-$ & $-$ & $-$ & $-$ & $-$ & $-$ & $-$ & $-$ & $-$ & $-$ & $-$ & $-$ & $-$ & Linear \\
		AutoCNN-S1-128 & 1 & 128 & 0 & 7x7x3 & $|x|$ & 4(4) & $-$ & $-$ & $-$ & $-$ & $-$ & $-$ & $-$ & $-$ & $-$ & $-$ & $-$ & $-$ & $-$ & Linear \\
		\vspace{3pt}
		AutoCNN-S1-128 + RA & 1 & 128 & 1-3 & 11x11x3 & $|x|$ & 4(4) & $-$ & $-$ & $-$ & $-$ & $-$ & $-$ & $-$ & $-$ & $-$ & $-$ & $-$ & $-$ & $-$ & Linear \\
		AutoCNN-S2 & 2 & 128 & 1-3 & 7x7x3 & $|x|$ & 5(3) & 1 & 1024 & 0-2 & 5x5x128 & $|x|$ & 3(2) & $-$ & $-$ & $-$ & $-$ & $-$ & $-$ & $-$ & Linear \\
		\midrule
		\multicolumn{21}{c}{\bf CIFAR-10 and CIFAR-100} \\
		\midrule
		AutoCNN-S1 & 1 & $K_1$ & 0-3 & $s_1$x$s_1$x3 & relu & 8(8) & $-$ & $-$ & $-$ & $-$ & $-$ & $-$ & $-$ & $-$ & $-$ & $-$ & $-$ & $-$ & $-$ & Linear \\
		AutoCNN-S1-256 & 1 & 256 & 0 & 9x9x3 & relu & 8(8) & $-$ & $-$ & $-$ & $-$ & $-$ & $-$ & $-$ & $-$ & $-$ & $-$ & $-$ & $-$ & $+$ & Linear \\
		\vspace{3pt}
		AutoCNN-S1-256 + RA & 1 & 256 & 0-3 & 13x13x3 & relu & 8(8) & $-$ & $-$ & $-$ & $-$ & $-$ & $-$ & $-$ & $-$ & $-$ & $-$ & $-$ & $-$ & $+$ & Linear \\
		AutoCNN-S32 & 2 & 128 & 0-3 & 9x9x3 & relu & 3(2) & 32 & 64 & 0-2 & 7x7x4 & $|x|$ & 5(4) & $-$ & $-$ & $-$ & $-$ & $-$ & $-$ & $+$ & Linear \\
		\vspace{3pt}
		AutoCNN-M2 & 2 & 256 & 0-3 & 5x5x3 & relu & 3(2) & 1 & 1024 & 0-2 & 5x5x256 & $|x|$ & 5(4) & $-$ & $-$ & $-$ & $-$ & $-$ & $-$ & $+$ & Linear \\
		AutoCNN-S & 3 & 64 & 0-3 & 5x5x3 & relu & 3(2) & 1 & 128 & 0-3 & 5x5x64 & $|x|$ & 3(2) & 1 & 256 & 0-2 & 5x5x128 & $|x|$ & 3(2) & $+$ & Linear \\
		AutoCNN-M & 3 & 92 & 0-3 & 5x5x3 & relu & 3(2) & 1 & 256 & 0-3 & 5x5x92 & $|x|$ & 3(2) & 1 & 512 & 0-2 & 5x5x256 & $|x|$ & 3(2) & $+$ & Linear \\
		\vspace{3pt}
		AutoCNN-L & 3 & 256 & 0-3 & 5x5x3 & relu & 3(2) & 1 & 1024 & 0-3 & 5x5x256 & $|x|$ & 3(2) & 1 & 2048 & 0-2 & 5x5x1024 & $|x|$ & 3(2) & $+$ & Linear \\
		AutoCNN-L32 & 3 & 1024 & 0-3 & 5x5x3 & relu & 3(2) & 32 & 256 & 0-2 & 5x5x32 & $|x|$ & 3(2) & 32 & 1024 & 0-2 & 5x5x256 & $|x|$ & 3(2) & $+$ & RBF \\
		\midrule
		\multicolumn{21}{c}{\bf STL-10} \\
		\midrule
		\vspace{3pt}
		AutoCNN-S1 & 1 & $K_1$ & 0-3 & $s_1$x$s_1$x3 & relu & 20(20) & $-$ & $-$ & $-$ & $-$ & $-$ & $-$ & $-$ & $-$ & $-$ & $-$ & $-$ & $-$ & $-$ & Linear \\
		AutoCNN-S & 3 & 64 & 0-3 & 7x7x3 & relu & 5(4) & 1 & 128 & 0-3 & 5x5x64 & $|x|$ & 4(3) & 1 & 256 & 0-2 & 5x5x128 & $|x|$ & 3(2) & $+$ & Linear \\
		AutoCNN-M & 3 & 92 & 0-3 & 7x7x3 & relu & 5(4) & 1 & 256 & 0-3 & 5x5x92 & $|x|$ & 4(3) & 1 & 512 & 0-2 & 5x5x256 & $|x|$ & 3(2) & $+$ & Linear \\
		\vspace{3pt}
		AutoCNN-L & 3 & 256 & 0-3 & 7x7x3 & relu & 5(4) & 1 & 1024 & 0-3 & 5x5x256 & $|x|$ & 4(3) & 1 & 2048 & 0-2 & 5x5x1024 & $|x|$ & 3(2) & $+$ & Linear \\
		AutoCNN-L32 & 3 & 1024 & 0-3 & 7x7x3 & relu & 5(4) & 32 & 256 & 0-2 & 5x5x32 & $|x|$ & 4(3) & 32 & 1024 & 0-2 & 5x5x256 & $|x|$ & 3(2) & $+$ & RBF \\
		\bottomrule
	\end{tabular}
	
	\justify
	\small
	
	$L$ - number of layers; \\
	$K_l$ - number of filters in layer $l$; \\
	$s_l,d_l$ - filter size and depth for layer $l$; \\
	$n_l$ - recursive autoconvolution (RA) order to learn filters for layer $l$ (in case RA is not applied $n_l=0$); \\
	$m_l$ - max pooling size and stride (in parentheses); for STL-10 in case of data augmentation $m_l = 4(3)$ instead of 5(4) in the first layer; \\
	$G_l$ - number of feature map groups; \\
	$R_l$ - rectifier after layer $l$ (relu:  $\max(0,x)$, $|x|$ - absolute values); \\
	root - Rootsift normalization after each layer ($\mathrm{sign}(\mathbf{x})\sqrt{| \mathbf{x} |/ \| \mathbf{x} \|_1 }$). \\
\end{table*}
\vspace*{5in}

\end{document}